\pdfoutput=1

\documentclass[11pt]{article}

\usepackage[]{emnlp2021}

\usepackage{times}
\usepackage{latexsym}

\usepackage[T1]{fontenc}

\usepackage[utf8]{inputenc}

\usepackage{microtype}
\usepackage{amsmath}
\usepackage{amsfonts}
\usepackage{booktabs}
\usepackage{float}
\usepackage{newfloat}
\usepackage{listings}
\usepackage{multirow}
\usepackage{multicol}
\usepackage{graphicx}

\newcommand*\samethanks[1][\value{footnote}]{\footnotemark[#1]}

\title{Efficient Sub-structured Knowledge Distillation}

\author{Wenye Lin$^{1,2}$\thanks{\ \ Equal contributions.}, Yangming Li$^1$\samethanks, Lemao Liu$^1$, Shuming Shi$^1$, Hai-tao Zheng$^2$\thanks{$^{\dagger}$ Corresponding author.} \\
  $^1$Tencent AI Lab \\
  $^2$Tsinghua University \\
  \texttt{lwy20@mails.tsinghua.edu.cn,zheng.haitao@sz.tsinghua.edu.cn} \\
  \texttt{\{newmanli,redmondliu,shumingshi\}@tencent.com}}

\begin{document}
\maketitle
\begin{abstract}

    Structured prediction models aim at solving a type of problem where the output is a complex structure, rather than a single variable. Performing knowledge distillation for such models is not trivial due to their exponentially large output space. In this work, we propose an approach that is much simpler in its formulation and far more efficient for training than existing approaches. Specifically, we transfer the knowledge from a teacher model to its student model by locally matching their predictions on all sub-structures, instead of the whole output space. In this manner, we avoid adopting some time-consuming techniques like dynamic programming (DP) for decoding output structures, which permits parallel computation and makes the training process even faster in practice. Besides, it encourages the student model to better mimic the internal behavior of the teacher model. Experiments on two structured prediction tasks demonstrate that our approach outperforms previous methods and halves the time cost for one training epoch.\footnote{The source code for this work is publicly available at https://github.com/Linwenye/Efficient-KD.}
    

\end{abstract}

\section{Introduction}
\begin{figure*}
         \centering
         \includegraphics[width=1\linewidth]{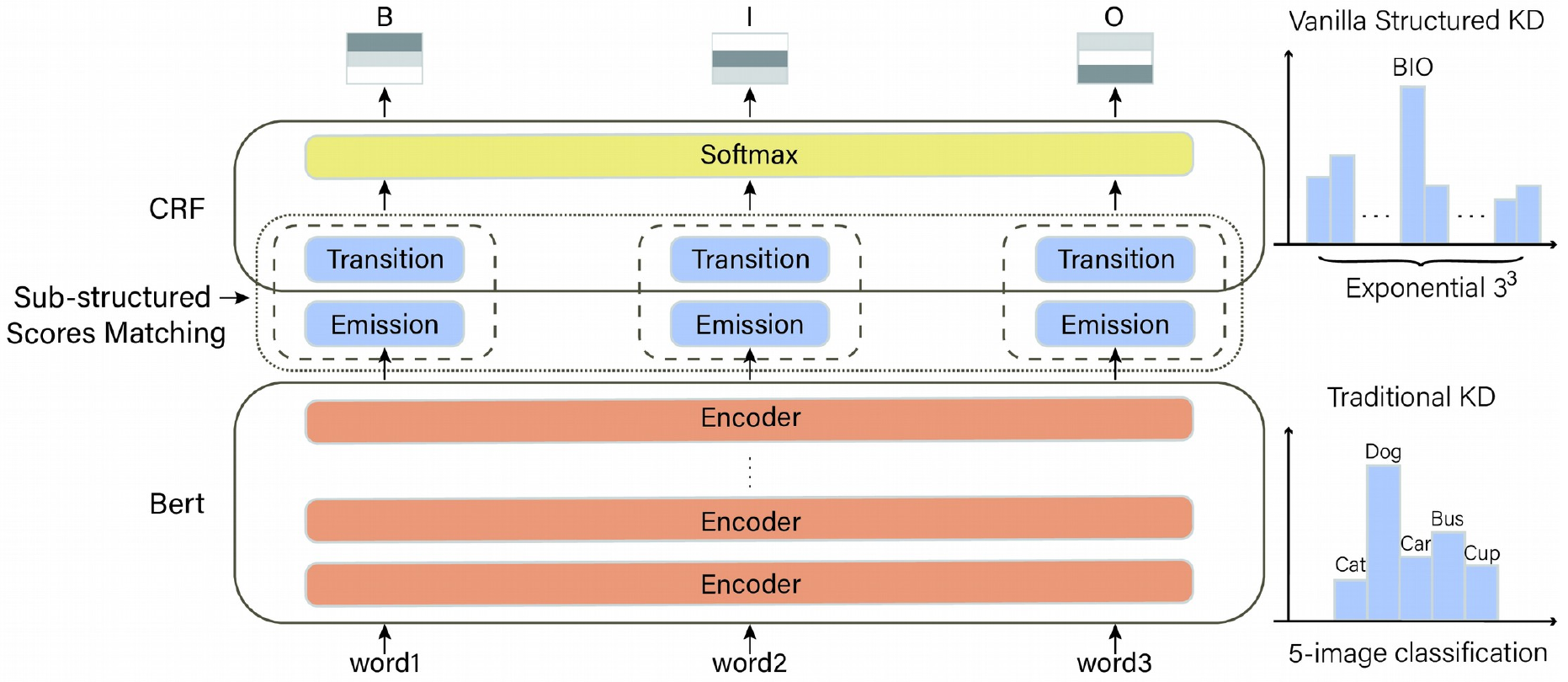}
         \caption{An example of our efficient sub-structured knowledge distillation. KD is short for knowledge distillation. Traditional KD matches output distributions directly. However, the output space of structured prediction models is exponential in size, and  thus infeasible to directly match output distributions. By matching sub-structured scores, we efficiently transfer the structured knowledge from the teacher to the student. }
         \label{image:kd}
\end{figure*}

    Large-scale neural networks have achieved great success in natural language processing (NLP) \citep{conneau-etal-2017-deep,devlin-etal-2019-bert,lewis-etal-2020-bart}. However, training and deploying these models are extremely computationally expensive. To improve their efficiency for applications, knowledge distillation is introduced by \citet{bucilu2006model,hinton2015distilling} to transfer the knowledge from a large teacher model to a small student model, by letting the student softly mimic the predictions of its teacher. 

    Prior works~\cite{sanh2019distilbert,sun-etal-2020-knowledge-distillation,li-etal-2021-dynamic} have presented a number of effective knowledge distillation approaches. However, these methods are inapplicable for structured prediction models that capture the mutual dependencies among output variables, because of their huge output space.
    Part-of-speech (POS) tagging~\citep{ratnaparkhi1996maximum} is a typical structured prediction task where two adjacent output tags are strongly correlated. For example, nouns usually follow adjectives, but not vice versa. To incorporate such correlations among tags, popular POS taggers such as CRF~\cite{lafferty2001conditional} make predictions at the sequence level, instead of over its single words. Since the size of the output space exponentially increases with the sequence length, directly applying traditional knowledge distillation is computationally infeasible.

    A small stream of recent works called structured knowledge distillation have studied this problem. \citet{kim_sequence-level_2016} approximate predictions of the teacher model by generating K-best outputs as pseudo labels. \citet{wang_structure-level_2020} compute a posterior marginal probability distribution for every output variable. \citet{wang_structural_2021} minimize the factorized form of the knowledge distillation loss. However, these approaches are very inefficient for training because their computations involve time-consuming operations like K-best decoding or dynamic programming (DP).

    In this work, we propose a highly efficient approach for structured knowledge distillation. Our idea is to locally match all sub-structured predictions of the teacher model and those of its student model, which avoids adopting time-consuming techniques like DP to globally search for output structures. Intuitively, similar predictions over the sub-structures lead to an accurate approximation of complete structures. Furthermore, mimicking the sub-structured predictions allows the student model to better imitate the internal behavior of the teacher model~\citep{ba2014deep}. For example, if a label sequences $[AA]$ is given score $10$ and $-10$ by the teacher model while its student predicts all 0 for both of $A$, the student learns nothing from the teacher if the globally structured knowledge distillation is adopted as the training objective since they all sum up to 0. Performing sub-structured knowledge distillation, however, enables the student model to learn the detailed behavior of the teacher.
    Besides, our simple formulation benefits in parallel matrix computation, which further reduces the training time.


    We have performed experiments on two structured prediction tasks, named entity recognition (NER) and syntactic chunking, showing that our approach outperforms previous methods with far less training time.

\section{Methodology}

    In this section, we provide a mathematical formulation of our approach. For better understanding, we assume that the model adopts CRF for decoding, which prevails in the literature of structured prediction. Our approach is in fact applicable for almost all structured prediction models.
    
\subsection{CRF}

    CRF is widely used in structured prediction tasks. For an input sequence $\mathbf{x}=\{x_1,x_2,...,x_L\}$ consisting of $L$ tokens, the corresponding ground truth is $\mathbf{y}=\{ y_1,y_2,...,y_L\}, y_l \in \{1,2,...,N\} $, where N is the size of the label set. To get expressive semantic representation for tokens, we adopt pre-trained BERT \cite{devlin2019bert}:
    \begin{equation}
        \mathbf{h}=\{\mathbf{h}_1,\mathbf{h}_2,...,\mathbf{h}_L\}=\rm{BERT}(\mathbf{x}).
    \end{equation}
    Then, we use a learnable matrix $\mathbf{W}$ to map token representation into emission scores:
    \begin{equation}
        \mathbf{e}=\{\mathbf{e}_1,\mathbf{e}_2,...,\mathbf{e}_L\}=\mathbf{W}\mathbf{h},
    \end{equation}
    where $\mathbf{e}_l$ is a $N$-dimensional vector. CRF considers not only the emission scores but also the transition scores $t_{y_{l-1}, y_l}$, so the score function for every sub-structure is
    \begin{equation}
        s\left(y_{l}, y_{l-1}, \mathbf{x}_{l}\right) = t_{y_{l-1},y_l} + \mathbf{e}_l.
    \end{equation}
    Given all sub-structure scores, the conditional probability $p(\mathbf{y}|\mathbf{x})$ takes the form:
    \begin{equation}
    \label{equ:p(y|x)}
    p(\mathbf{y}|\mathbf{x})=\frac{1}{Z(\mathbf{x})} \prod_{l=1}^{L} \exp \left\{s\left(y_{l}, y_{l-1}, \mathbf{x}_{l}\right)\right\},
    \end{equation}
    where $Z(\mathbf{x})$ is a normalization term computed by Viterbi algorithm.
    
\subsection{Standard Knowledge Distillation}

    Knowledge distillation is a ubiquitous technique for model compression and acceleration. It trains a small student model to imitate the output distribution of a large teacher model via $L_2$ distance on logits~\cite{ba2014deep} or cross-entropy~\cite{hinton2015distilling}: 
    \begin{equation}
    \label{kd}
        \mathcal{L}_{\mathrm{KD}}=-\sum_{\mathbf{y} \in Y(\mathbf{x})} P_{t}(\mathbf{y} | \mathbf{x}) \log P_{s}(\mathbf{y} | \mathbf{x}),
    \end{equation}
    where $P_t$ and $P_s$ are respectively the distributions of the teacher model and the student model.

    In practice, the overall training objective contains the above distillation loss and a negative log likelihood loss on the ground truth.
    \begin{equation}
        \mathcal{L}_{\mathrm {student }}=\lambda \mathcal{L}_{\mathrm{KD}}+(1-\lambda) \mathcal{L}_{\mathrm{NLL }},
    \end{equation}
    where $\lambda$ is usually set as a dynamic weight that starts from $1$ to $0$ during training.

\subsection{Structured Knowledge Distillation}

    Conventional knowledge distillation, i.e., Eq.~(\ref{kd}) is intractable for structured prediction models, since the output space $Y(\mathbf{x})$ grows exponentially with the sentence length $L$. To tackle this problem, \citet{kim_sequence-level_2016} utilize K-best sequences predicted by the teacher model to approximate the teacher distribution $P_t$:
    \begin{equation}
    p_{t}(\mathbf{y} | \mathbf{x})= \begin{cases}
    \frac{p_{t}(\mathbf{y} | \mathbf{x})}{\sum_{{\mathbf{y}} \in \mathcal{T}} p_{t}({\mathbf{y}} | \mathbf{x})} & \mathbf{y} \in \mathcal{T} \\ 0 & \mathbf{y} \notin \mathcal{T}\end{cases},
    \end{equation}
    where $\mathcal{T}$ is the set of predicted sequences. \citet{wang_structural_2021} calculate the factorized form of the Eq.~(\ref{kd}) for CRF model as
    \begin{equation}
        \mathcal{L}_{\mathrm{KD}} = -\sum_{\mathbf{u} \in \mathbb{U}(\mathbf{x})} p'(\mathbf{u}|\mathbf{x}) s(\mathbf{u}, \mathbf{x})+\log Z(\mathbf{x}),
    \end{equation}
    where $\mathbf{u}$ denotes any possible two adjacent tags $\{y_l,y_{l-1}\}$ and $p'$ is a conditional probability computed by the teacher model.
    
    While these methods allow tractable knowledge distillation, they introduce sequential calculations with high time complexity and are not parallelizable.
    To boost the efficiency, we propose to let the student model learn from its teacher model by locally matching their predictions on all sub-structures. Specifically, we adopt $L_2$ distance to measure the sub-structure prediction difference between the two models:
    \begin{equation}
    \label{equ:kd}
        \mathcal{L}_{\mathrm{KD}} = \frac{\sum_{\mathbf{u} \in \mathbb{U}(\mathbf{x})} \left \| \rm{s} \left ( \mathbf{u}, \mathbf{x} \right ) - \rm{s}' \left ( \mathbf{u}, \mathbf{x} \right )\right \|_2}{\left | \mathbb{U}(\mathbf{x}) \right |},
    \end{equation}
    where $s'$ represents sub-structure scores calculated by the teacher model and $\mathbb{U}$ denotes all possible pairs of adjacent tags.
\begin{table*}
    \caption{Comparison of precision, recall and f1 score results on test sets. All metrics are calculated using the CoNLL 2000 evaluation script. Reported values are the average of 5 runs with different random seeds. }
    \label{table:f1}
  \centering
    \begin{tabular}{lcccccc}
    \toprule
    \multirow{2}{*}[-2pt]{Method}& \multicolumn{3}{c}{CoNLL 2003 NER} & \multicolumn{3}{c}{CoNLL 2000 Chunking}
    \\\cmidrule(lr){2-4}\cmidrule(lr){5-7}
    &Precision & Recall & F1 & Precision & Recall & F1\\
    \midrule
    BERT$\rm{_{BASE}}$-CRF (teacher)    & 92.00 &  92.62  & 92.31      & 97.80        & 97.80& 97.80 \\
    BERT$\rm{_{3L}}$-CRF (student)    & 88.93 &  88.79   & 88.86      & 96.20   & 96.22 & 96.21 \\
    \midrule
    K-best \cite{kim_sequence-level_2016} & 88.94 &89.50 &89.22 & 96.46 & 96.41 & 96.44\\
    Struct. KD \cite{wang_structural_2021} &89.33&89.33&89.33&96.48&96.43&96.46\\
    \midrule
    Efficient KD &\textbf{89.69}&\textbf{89.71}&\textbf{89.70}&\textbf{96.81}&\textbf{96.77}&\textbf{96.79}\\
    \bottomrule

    \end{tabular}
\end{table*}

\begin{table*}
    \caption{The efficiency comparisons among baselines and our approach. We tabulate the time costs of different computation processes in one training epoch. Tea. and Stu. represent teacher and student respectively.}
    \label{tab:eff}

    \centering
    \begin{tabular}{lccccc}
    \toprule
     Method  & Tea. Forward & Stu. Forward & Stu. Backward & Total\\
     \midrule
      Vanilla Training & - & 19s & 35s& 54s\\
      \midrule
      K-best  & 21s & 23s & 39s & 83s \\
      Struct. KD  & 19s  & 30s & 50s & 99s \\
      Efficient KD & \textbf{16s}  & \textbf{5s}&\textbf{20s} &\textbf{41s}\\
      \bottomrule

    \end{tabular}

\end{table*}
\section{Experiments}


\subsection{Settings}

    We use CoNLL 2003 dataset~\cite{sang2003introduction} for experiments on NER and CoNLL 2000 dataset on text chunking. We follow standard training/development/test splits.

    \paragraph{Structured Prediction Model} For all structured prediction tasks, we choose the widely used BERT-CRF model, in which pre-trained BERT \cite{devlin2019bert} fully exploits the knowledge from massive unlabeled corpora and CRF captures the interrelationship of predicted variables. For the teacher model BERT$_{\rm{BASE}}$-CRF, we first load the pre-trained BERT$_{\rm{BASE}}$ cased version that consists of 12 Transformer Encoder layers and then finetune it on target task along with the CRF layer. For the student model BERT$\rm{_{3L}}$-CRF, since a pre-trained BERT$\rm{_{3L}}$ model is not directly available, we load the first three layers of pre-trained weights from the BERT$_{\rm{BASE}}$ model.
    
    \paragraph{Training} We conduct all experiments on one NVIDIA Tesla V100 GPU device. We set the batch size as 32 and initial learning rate as  $5\times 10^{-5}$ for the teacher model and $2\times 10^{-4}$ for the student model. We utilize Adam~\citep{kingma2014adam} as the optimizer.
    To perform structured knowledge distillation, we follow the following procedure: (1) Load the BERT$_{\rm{BASE}}$-CRF teacher model and finetune it on the target task; (2) Calculate sub-structure scores of the teacher model on the target dataset; (3) Train the student model to minimize the $L_2$ distance of sub-structure scores between the student model and the teacher model using Eq.~(\ref{equ:kd}). Note that for our efficient knowledge distillation method, $\lambda$ is set as 1 to fully utilize the parallelism, and is set to 0 only at the last epoch. This strategy improves the efficiency of our method without hurting its performance, due to our observation that the teacher model perfectly fits the training set.
        
    \paragraph{Baselines} All baselines of structured knowledge distillation \cite{kim_sequence-level_2016,wang_structural_2021,wang_structure-level_2020} are re-implemented with BERT-CRF model following the official implementations\footnote{https://github.com/Alibaba-NLP/MultilangStructureKD and https://github.com/Alibaba-NLP/StructuralKD} for fair comparisons. For K-best decoding, we empirically find that a K value larger than 3 does not further improve the performance but introduces more training cost, so K is set to be 3 by default. 

\subsection{Main Results}

    Table \ref{table:f1} shows the F1 scores of baselines and our approach on NER and syntactic chunking. Although we are focusing on improving the efficiency, we observe that our proposed approach (i.e., Efficient KD) performs well and even outperforms other existing structured knowledge distillation methods. This shows that imitating the sub-structure predictions indeed helps the student model better mimic the internal behavior of the teacher model. More detailedly, our observations are as follows: (1) Thanks to the informational representation from BERT plus the ability to construct the relationship of outputs from CRF, our teacher model BERT$\rm{_{BASE}}$-CRF reaches new state-of-the-art performance (i.e., 97.80) on CoNLL 2000 chunking task; (2) struct. KD \cite{wang_structural_2021} performs slightly better than the K-best method \cite{kim_sequence-level_2016} for structured prediction tasks, while they all boost the performance of the student model; (3) our proposed method surpasses previous baselines and improves the performance of the student model by a significant margin.

    \subsection{Efficiency Analysis}
    To demonstrate that our method distills the structured knowledge more efficiently, we measure the training speed for different distillation methods on the CoNLL 2003 dataset per epoch. The batch size is 32 and the running device is a NVIDIA Tesla V100 GPU. The running time is calculated as the average of 5 epochs after warmup of GPU. The results are shown in Table \ref{tab:eff}. It shows that our proposed method significantly speeds up the training process, which is composed of three parts: teacher forward, student forward and student backward. We analyze the reasons as follows: The forward speed of the teacher model is slightly quicker simply because the calculation of sub-structure scores does not include the calculation in Eq.~(\ref{equ:p(y|x)}). The time cost of the student forward and backward is largely reduced given the credit to the computational parallelism of our method. Longer sequence, more speed-up. As a result, our method costs less than half of the training time for the student model (41s compared to 83s and 99s one epoch). These results verify the efficiency of our method. 

\section{Conclusion}

    While a few recent works have presented tractable approaches for structured knowledge distillation, they are time-consuming due to sequence-level computation. In this work, we propose to locally match the predictions of the student model and the teacher model on all sub-structures. In this way, we avoid globally searching for output structures and help the student model better mimic the internal behavior of its teacher. Besides, the simple formulation of our approach allows parallel matrix computation on GPUs. We conduct experiments on NER and text chunking across two datasets, showing that our approach significantly outperforms previous baselines and owns very high training efficiency.

\bibliography{custom}
\bibliographystyle{acl_natbib}




\end{document}